\documentclass[letterpaper]{sig-alternate-05-2015}
\usepackage{amsfonts}
\usepackage[colorlinks, citecolor=blue, urlcolor=cyan, linkcolor=red]{hyperref}
\usepackage{amsmath,amssymb,amscd,epsfig,amsfonts,rotating}
\usepackage{graphicx}
\usepackage{epstopdf}
\usepackage{subfigure}
\usepackage{multirow}
\usepackage{booktabs}
\usepackage{color,xcolor}
\usepackage{url}
\usepackage{amsmath,amscd,amsbsy,amssymb,latexsym,url,bm}
\usepackage{epsfig,epstopdf,graphicx,subfigure}
\usepackage{enumitem,balance,mathtools}
\usepackage{wrapfig}
\usepackage{mathrsfs, euscript}
\usepackage{algorithm}
\usepackage{algorithmic}
\usepackage{amssymb}
\usepackage{ifpdf}
\usepackage{multicol}
\DeclareMathOperator*{\argmax}{argmax}

\newcommand{\bs}{\boldsymbol}
\newcommand{\bx}{\bs{x}}

\newcommand{\tsc}{\textsc}

\newcommand{\minisection}[1]{\vspace{5pt}\noindent\textbf{#1.}}

\begin{document}

\CopyrightYear{2017}
\setcopyright{acmcopyright}
\conferenceinfo{WSDM 2017,}{February 06-10, 2017, Cambridge, United Kingdom}
\isbn{978-1-4503-4675-7/17/02}\acmPrice{\$15.00}
\doi{http://dx.doi.org/10.1145/3018661.3018702}

\clubpenalty=10000
\widowpenalty = 10000

\title{Real-Time Bidding by Reinforcement Learning\\in Display Advertising}

\numberofauthors{1}
\author{
	\alignauthor
	$^\dag$Han Cai, $^\dag$Kan Ren, $^\dag$Weinan Zhang,\thanks{Weinan Zhang is the corresponding author of this paper.} \\$^\ddag$Kleanthis Malialis, $^{\ddag \natural}$Jun Wang, $^\dag$Yong Yu, $^\sharp$Defeng Guo\\
	\affaddr{$^\dag$Shanghai Jiao Tong University, $^\ddag$University College London, $^\natural$MediaGamma Ltd, $^\sharp$Vlion Inc.}\\
	\email{\{hcai,kren,wnzhang\}@apex.sjtu.edu.cn, j.wang@cs.ucl.ac.uk}
}

\maketitle

\begin{abstract}
The majority of online display ads are served through real-time bidding (RTB) --- each ad display impression is auctioned off in real-time when it is just being generated from a user visit. To place an ad automatically and optimally, it is critical for advertisers to devise a learning algorithm to cleverly bid an ad impression in real-time. Most previous works consider the bid decision as a static optimization problem of either treating the value of each impression independently or setting a bid price to each segment of ad volume. However, the bidding for a given ad campaign would repeatedly happen during its life span before the budget runs out. As such, each bid is strategically correlated by the constrained budget and the overall effectiveness of the campaign (e.g., the rewards from generated clicks), which is only observed after the campaign has completed.  Thus, it is of great interest to devise an optimal bidding strategy sequentially so that the campaign budget can be dynamically allocated across all the available impressions on the basis of both the immediate and future rewards.
In this paper, we formulate the bid decision process as a reinforcement learning problem, where the state space is represented by the auction information and the campaign's real-time parameters, while an action is the bid price to set. By modeling the state transition via auction competition, we build a Markov Decision Process framework for learning the optimal bidding policy to optimize the advertising performance in the dynamic real-time bidding environment. Furthermore, the scalability problem from the large real-world auction volume and campaign budget is well handled by state value approximation using neural networks.
The empirical study on two large-scale real-world datasets and the live A/B testing on a commercial platform have demonstrated the superior performance and high efficiency compared to state-of-the-art methods.
\end{abstract}
\vspace{-5pt}

\section{Introduction}\label{sec:intro}
The increased availability of big data and the improved computational power have advanced machine learning and artificial intelligence for various prediction and decision making tasks. In particular, the successful application of reinforcement learning in certain settings such as gaming control \cite{silver2016mastering} has demonstrated that machines not only can predict, but also have a potential of achieving comparable human-level control and decision making. In this paper, we study \emph{machine bidding} in the context of display advertising. Auctions, particularly real-time bidding (RTB),  have been a major trading mechanism for online display advertising \cite{yuan2013real,google2011arrival,wang2015real}. Unlike the keyword-level bid decision in sponsored search \cite{amin2012budget}, the advertiser needs to make the impression-level bid decision in RTB, i.e., bidding for every single ad impression in real time when it is just being generated by a user visit \cite{perlich2012bid,yuan2013real} . Machine based bid decision, i.e., to calculate the strategic amount that the advertiser would like to pay for an ad opportunity, constitutes a core component that drives the campaigns' ROI \cite{hosanagar2008optimal,zhang2014optimal}. By calculating an optimal bid price for each ad auction (also considering the remaining budget and the future availability of relevant ad impressions in the ad exchange) and then observing the auction result and user feedback, the advertiser would be able to refine their bidding strategy and better allocate the campaign budget across the online page view volume.


A straightforward bidding strategy in RTB display advertising is mainly based on the truthfulness of second-price auctions \cite{edelman2005internet}, which means the bid price for each ad impression should be equal to its true value, i.e., the action value (e.g., click value) multiplied by the action rate (e.g., click-through rate) \cite{lee2012estimating}. However, for budgeted bidding in repeated auctions, the optimal bidding strategy may not be truthful but depends on the market competition, auction volume and campaign budget \cite{zhang2015statistical}. In \cite{perlich2012bid,zhang2014optimal}, researchers have proposed to seek the optimal bidding function that directly maximizes the campaign's key performance indicator (KPI), e.g., total clicks or revenue, based on the static distribution of input data and market competition models. Nevertheless, such static bid optimization frameworks may still not work well in practice because the RTB market competition is highly dynamic and it is fairly common that the true data distribution heavily deviates from the assumed one during the model training \cite{chen2011real}, which requires additional control step such as budget pacing \cite{xu2015smart} to constrain the budget spending.

In this paper, we solve the issue by considering bidding as a sequential decision, and formulate it as a \emph{reinforcement learning to bid} (RLB) problem. From an advertiser's perspective, the whole ad market and Internet users are regarded as the environment. At each timestep, the advertiser's bidding agent observes a state, which consists of the campaign's current parameters, such as the remaining lifetime and budget, and the bid request for a specific ad impression (containing the information about the underlying user and their context) . With such state (and context), the bidding agent makes a bid action for its ad. After the ad auction, the winning results with the cost and the corresponding user feedback will be sent back to the bidding agent, which forms the reward signal of the bidding action. Thus, the bid decision aims to derive an optimal bidding policy for each given bid request.


With the above settings, we build a Markov Decision Process (MDP) framework for learning the optimal bidding policy to optimize the advertising performance. The value of each state will be calculated by performing dynamic programming. Furthermore, to handle the scalability problem for the real-world auction volume and campaign budget, we propose to leverage a neural network model to approximate the value function. Besides directly generalizing the neural network value function, we also propose a novel \emph{coarse-to-fine episode segmentation model} and \emph{state mapping models} to overcome the large-scale state generalization problem.

In our empirical study, the proposed solution has achieved 16.7\% and 7.4\% performance gains against the state-of-the-art methods on two large-scale real-world datasets. In addition, our proposed system has been deployed into a commercial RTB platform. We have performed an online A/B testing, where a 44.7\% improvement in click performance was observed against a most widely used method in the industry.

\section{Background and Related Work}\label{sec:related}
\minisection{Reinforcement Learning} An MDP provides a mathematical framework which is widely used for modelling the dynamics of an environment under different actions, and is useful for solving reinforcement learning problems \cite{sutton1998introduction}. An MDP is defined by the tuple $\langle S,A,P,R \rangle$. The set of all states and actions are represented by $S$ and $A$ respectively. The reward and transition probability functions are given by $R$ and $P$. Dynamic programming is used in cases where the environment's dynamics, i.e., the reward function and transition probabilities are known in advance. Two popular dynamic programming algorithms are policy iteration and value iteration. For large-scale situations, it is difficult to experience the whole state space, which leads to the use of function approximation that constructs an approximator of the entire function \cite{gordon1995stable,taylor2009kernelized}. In this work, we use value iteration for small-scale situations, and further build a neural network approximator to solve the scalability problem.

\minisection{RTB Strategy} In the RTB process \cite{zhang2014optimal}, the advertiser receives the bid request of an ad impression with its real-time information and the very first thing to do is to estimate the \emph{utility}, i.e., the user's response on the ad if winning the auction.
The distribution of the \emph{cost}, i.e., the market price \cite{amin2012budget,cui2011bid}, which is the highest bid price from other competitors, is also forecasted by the \emph{bid landscape forecasting} component. Utility estimation and bid landscape forecasting are described below.
Given the estimated utility and cost factors, the \emph{bidding strategy} \cite{yuan2013real} decides the final bid price with accessing the information of the remaining budget and auction volume.
Thus it is crucial to optimize the final bidding strategy considering the market and bid request information with budget constraints.
A recent comprehensive study on the data science of RTB display advertising is posted in \cite{wang2016display}. 

\minisection{Utility Estimation} For advertisers, the utility is usually defined based on the user response, i.e., click or conversion, and can be modeled as a probability estimation task \cite{lee2012estimating}. Much work has been proposed for user response prediction, e.g., click-through rate (CTR) \cite{mcmahan2013ad}, conversion rate (CVR) \cite{lee2012estimating} and post-click conversion delay patterns \cite{chapelle2014modeling}. For modeling, linear models such as logistic regression \cite{lee2012estimating} and non-linear models such as boosted trees \cite{he2014practical} and factorization machines \cite{oentaryo2014predicting} are widely used in practice. There are also online learning models that immediately perform updating when observing each data instance, such as Bayesian probit regression \cite{graepel2010web}, FTRL learning in logistic regression \cite{mcmahan2013ad}. In our paper, we follow \cite{lee2012estimating,zhang2014optimal} and adopt the most widely used logistic regression for utility estimation to model the reward on agent actions.


\minisection{Bid Landscape Forecasting} Bid landscape forecasting refers to modeling the market price distribution for auctions of specific ad inventory, and its c.d.f. is the winning probability given each specific bid price \cite{cui2011bid}.
The authors in \cite{li2014programmatic,zhang2014optimal,cui2011bid} presented some hypothetical winning functions and learned the parameters. For example, a log-normal market price distribution with the parameters estimated by gradient boosting decision trees was proposed in \cite{cui2011bid}.
Since advertisers only observe the winning impressions, the problem of censored data \cite{amin2012budget,zhang2016bid} is critical. Authors in \cite{wu2015predicting} proposed to leverage censored linear regression to jointly model the likelihood of observed market prices in winning cases and censored ones with losing bids.
Recently, the authors in \cite{wang2016functional} proposed to combine survival analysis and decision tree models, where each tree leaf maintains a non-parametric survival model to fit the censored market prices.
In this paper, we follow \cite{amin2012budget,zhang2016bid} and use a non-parametric method to model the market price distribution.

\minisection{Bid Optimization} As has been discussed above, bidding strategy optimization is the key component within the decision process for the advertisers \cite{perlich2012bid}. The auction theory \cite{krishna2009auction} proved that truthful bidding is the optimal strategy under the second-price auction. However, truthful bidding may perform poorly when considering the multiple auctions and the budget constraint \cite{zhang2014optimal}. In real-world applications, the linear bidding function \cite{perlich2012bid} is widely used. The authors in \cite{zhang2014optimal} empirically showed that there existed non-linear bidding functions better than the linear ones under variant budget constraints. When the data changes, however, the heuristic model \cite{perlich2012bid} or hypothetical bidding functions \cite{zhang2015statistical,zhang2014optimal} cannot depict well the real data distribution. The authors in \cite{amin2012budget,yuan2012sequential} proposed the model-based MDPs to derive the optimal policy for bidding in sponsored search or ad selection in contextual advertising, where the decision is made on keyword level.
In our work, we investigate the most challenging impression-level bid decision problem in RTB display advertising that is totally different from \cite{amin2012budget,yuan2012sequential}. We also tackle the scalability problem, which remains unsolved in \cite{amin2012budget}, and demonstrate the efficiency and effectiveness of our method in a variety of experiments.

\section{Problem and Formulation}\label{sec:method}

\begin{figure}[t]
	\centering
	\includegraphics[width=\columnwidth]{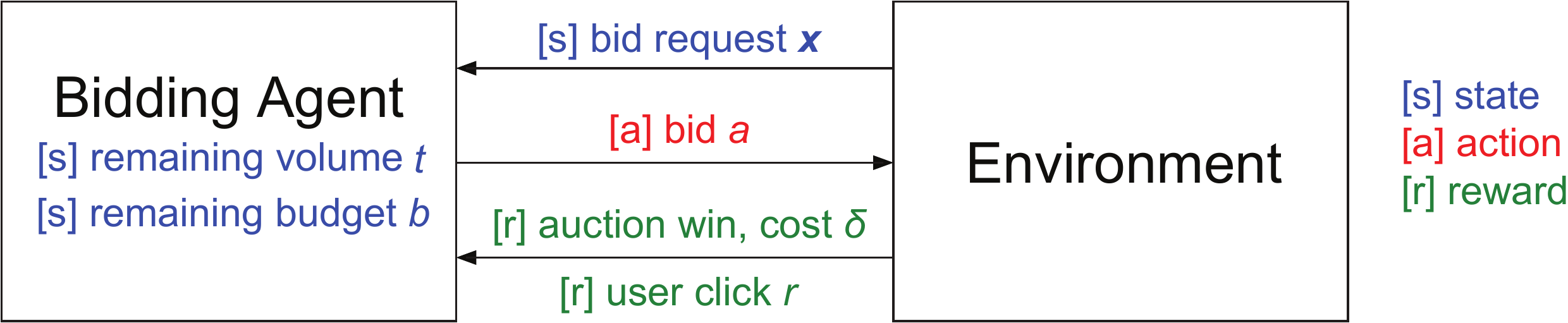}
	\caption{Diagram of reinforcement learning to bid.}\label{fig:rl-bid}
\end{figure}

In a RTB ad auction, each bidding agent acts on behalf of its advertiser and generates bids to achieve the campaign's specific target. Our main goal is to derive the optimal bidding policy in a reinforcement learning fashion. For most performance-driven campaigns, the optimization target is to maximize the user responses on the displayed ads if the bid leads to auction winning.
Without loss of generality, we consider clicks as our targeted user response objective, while other KPIs can be adopted similarly.
The diagram of interactions between the bidding agent and the environment is shown in Figure~\ref{fig:rl-bid}.

\subsection{Problem Definition}

Mathematically, we consider bidding in display advertising as an episodic process \cite{amin2012budget}; each episode comprises $T$ auctions which are sequentially sent to the bidding agent.
Each auction is represented by a high dimensional feature vector $\bx$, which is indexed via one-hot binary encoding. Each entry of $\bx$ corresponds to a category in a field, such as the category \texttt{London} in the field \texttt{City}, and the category \texttt{Friday} in the field \texttt{Weekday}. The fields consist of the campaign's ad information (e.g., ad creative ID and campaign ID) and the auctioned impression contextual information (e.g., user cookie ID, location, time, publisher domain and URL).

At the beginning, the agent is initialized with a budget $B$, and the advertising target is set to acquire as many clicks as possible during the following $T$ auctions.
Three main pieces of information are considered by the agent (i) the remaining auction number $t \in \{0,\cdots,T\}$; (ii) the unspent budget $b \in \{0,\cdots,B\}$ and (iii) the feature vector $\bx$.
During the episode, each auction will be sent to the agent sequentially and for each of them the agent needs to decide the bid price according to the current information $t$, $b$ and $\bx$.

The agent maintains the remaining number of auctions $t$ and the remaining budget $b$.
At each timestep, the agent receives an auction $\bx \in \bs{X}$ (the feature vector space), and determines its bid price $a$. We denote the market price p.d.f. given $\bx$ as $m(\delta, \bx)$, where $\delta$ is the market price and $m$ is its probability. If the agent bids at price $a \geq \delta$, then it wins the auction and pays $\delta$, and the remaining budget changes to $b - \delta$. In this case, the agent can observe the user response and the market price later. Alternatively, if losing, the agent gets nothing from the auction.
We take predicted CTR (pCTR) $\theta(\bx)$ as the expected reward, to model the action utility.
After each auction, the remaining number of auctions changes to $t - 1$.
When the auction flow of this episode runs out, the current episode ends and both the remaining auction number and budget are reset.

\subsection{MDP Formulation of RTB}
A Markov Decision Process (MDP) provides a framework that is widely used for modeling agent-environment interactions \cite{sutton1998introduction}.
Our notations are listed in Table~\ref{tab:notation-description}.
An MDP can be represented by a tuple $(\bs{S}, \{\bs{A}_s\}, \{P_{ss'}^a\}, \{R_{ss'}^a\})$, where $\bs{S}$ and $\bs{A}_s$ are two sets of all states and all possible actions in state $s \in \bs{S}$, respectively. $P_{ss'}^a$ represents the state transition probability from state $s \in \bs{S}$ to another state $s' \in \bs{S}$ when taking action $a \in \bs{A}_s$, which is denoted by $\mu(a, s, s')$.
Similarly, $R_{ss'}^a$ is the reward function denoted by $r(a, s, s')$ that represents the reward received after taking action $a$ in state $s$ and then transiting to state $s'$.

\begin{table}[t]
\centering
\caption{A summary of our notations.}
\label{tab:notation-description}
\resizebox{\columnwidth}{!}{
\begin{tabular}{c | l}
	Notation & Description \\
	\hline
	$\bx$ & The feature vector that represents a bid request. \\
	$\bs{X}$ & The whole feature vector space. \\
	$p_x(\bx)$ & The probability density function of $\bx$. \\
	$\theta(\bx)$ & The predicted CTR (pCTR) if winning the auction of $\bx$. \\
	$m(\delta, \bx)$ & The p.d.f. of market price $\delta$ given $\bx$. \\
	$m(\delta)$ & The p.d.f. of market price $\delta$. \\
	$V(t, b, \bx)$ & The expected total reward with starting state $(t, b, \bx)$, \\
	& taking the optimal policy. \\
	$V(t, b)$ & The expected total reward with starting state $(t, b)$, \\
	& taking the optimal policy. \\
	$a(t, b, \bx)$ & The optimal action in state $(t, b, \bx)$. \\
\end{tabular}
}
\end{table}

We consider $(t, b, \bx_t)$ as a state $s$\footnote{For simplicity, we slightly abuse the notation by including $t$ in the state.} and assume the feature vector $\bx_t$ is drawn i.i.d. from the probability density function $p_x(\bx)$.
The full state space is $\bs{S} = \{0,\cdots,T\} \times \{0,\cdots,B\} \times \bs{X}$.
And if $t = 0$, the state is regarded as a terminal state which means the end of the episode.
The set of all actions available in state $(t, b, \bx_t)$ is $\bs{A}_{(t, b, \bx_t)} = \{0,\cdots,b\}$, corresponding to the bid price.
Furthermore, in state $(t, b, \bx_t)$ where $t > 0$, the agent, when bidding $a$, can transit to $(t - 1, b - \delta, \bx_{t-1})$ with probability $p_x(\bx_{t-1}) m(\delta, \bx_t)$ where $\delta \in \{0,\cdots,a\}$ and $\bx_{t-1} \in \bs{X}$. That is the case of winning the auction and receiving a reward $\theta(\bx_t)$.
And the agent may lose the auction whereafter transit to $(t - 1, b, \bx_{t-1})$ with probability $p_x(\bx_{t-1}) \sum_{\delta = a+1}^\infty m(\delta, \bx_t)$, where $\bx_{t-1} \in \bs{X}$.
All other transitions are impossible because of the auction process.
In summary, transition probabilities and reward function can be written as:
{\small
\begin{align}
\label{eq:transition-probability&reward-function}
	\mu\Big(a, (t, b, \bx_t), (t - 1, b-\delta, \bx_{t-1})\Big) &= p_x(\bx_{t-1}) m(\delta, \bx_t), \nonumber\\
	\mu\Big(a, (t, b, \bx_t), (t-1, b, \bx_{t-1})\Big) &= p_x(\bx_{t-1}) \sum_{\delta = a+1}^\infty m(\delta, \bx_t), \nonumber\\
	r\Big(a, (t, b, \bx_t), (t-1, b-\delta, \bx_{t-1})\Big) &= \theta(\bx_t), \nonumber\\
	r\Big(a, (t, b, \bx_t), (t-1, b, \bx_{t-1})\Big) &= 0,
\end{align}
}where $\delta \in \{0,\cdots,a\}$, $\bx_{t-1} \in \bs{X}$ and $t > 0$. Specifically, the first equation is the transition when giving a bid price $a \geq \delta$, while the second equation is the transition when losing the auction.


A deterministic policy $\pi$ is a mapping from each state $s \in \bs{S}$ to action $a \in \bs{A}_s$, i.e., $a = \pi(s)$, which corresponds to the bidding strategy in RTB display advertising.
According to the policy $\pi$, we have the value function $V^\pi(s)$: the expected sum of rewards upon starting in state $s$ and obeying policy $\pi$.
This satisfies the Bellman equation with the discount factor $\gamma = 1$ since in our scenario the total click number is the optimization target, regardless of the click time.
\begin{equation}
\small
	V^\pi(s) = \sum_{s' \in \bs{S}} \mu(\pi(s), s, s') \Big( r(\pi(s), s, s') + V^\pi(s') \Big)
\end{equation}
The optimal value function is defined as $V^*(s) = \max_{\pi} V^\pi(s)$. We also have the optimal policy as:
\begin{equation}
\label{eq:policy-func}
\small
	\pi^*(s) = \argmax_{a \in \bs{A}_s} \Big\{\sum_{s' \in \bs{S}} \mu(a, s, s') \Big(r(a, s, s') + V^*(s')\Big)
    \Big\},
\end{equation}
which gives the optimal action at each state $s$ and $V^*(s) = V^{\pi^*}(s)$.
The optimal policy $\pi^*(s)$ is exactly the optimal bidding strategy we want to find. For notation simplicity, in later sections, we use $V(s)$ to represent the optimal value function $V^*(s)$.

One may consider the possibility of model-free approaches \cite{watkins1992q,strehl2006pac} to directly learn the bidding policy from experience. However, such model-free approaches may suffer from the problems of transition dynamics of the enormous state space, the sparsity of the reward signals and the highly stochastic environment. Since there are many previous works on modeling the utility (reward) and the market price distribution (transition probability) as discussed in Section~\ref{sec:related}, we take advantage of them and propose our model-based solution for this problem.



\section{Dynamic Programming Solution}
In a small scale, Eq.~(\ref{eq:policy-func}) can be solved using a dynamic programming approach.
As defined, we have the optimal value function $V(t, b, \bx)$, where $(t, b, \bx)$ represents the state $s$. Meanwhile, we consider the situations where we do not observe the feature vector $\bx$; so another optimal value function is $V(t, b)$:
the expected total reward upon starting in $(t, b)$ without observing the feature vector $\bx$ when the agent takes the optimal policy. It satisfies
$V(t, b) = \int_{\bs{X}} p_x(\bx)$ $V(t, b, \bx) \,d{\bx}$.
Also that, we have the optimal policy $\pi^*$ and express it as the optimal action $a(t, b, \bx)$.


From the definition, we have $V(0, b, \bx) = V(0, b) = 0$ as the agent gets nothing when there are no remaining auctions. Combined with the transition probability and reward function described in Eq.~(\ref{eq:transition-probability&reward-function}), the definition of $V(t, b)$,  $V(t, b, \bx)$ can be expressed with $V(t - 1, \cdot)$ as
{\small
\begin{align}
\label{eq:value-function-relation}
\small
	V(t, b, \bx) &= \max_{0 \leq a \leq b} \Big\{
	\sum_{\delta = 0}^a \int_{\bs{X}}
	m(\delta, \bx) p_x(\bx_{t-1})~\cdot \nonumber\\
	&~~~~\Big(\theta(\bx) + V(t-1, b-\delta, \bx_{t-1})\Big) \,d{\bx_{t-1}}~+ \nonumber\\
	&~~~\sum_{\delta = a+1}^\infty \int_{\bs{X}}
	m(\delta, \bx) p_x(\bx_{t-1}) V(t-1, b, \bx_{t-1}) \,d{\bx_{t-1}}
	\Big\} \nonumber\\
	&= \max_{0 \leq a \leq b} \Big\{
	\sum_{\delta = 0}^a m(\delta, \bx)\Big(\theta(\bx) + V(t-1, b-\delta)\Big)~+ \nonumber\\
	&~~~\sum_{\delta = a+1}^\infty m(\delta, \bx) V(t-1, b)
	\Big\},
\end{align}
}where the first summation\footnote{In practice, the bid prices in various RTB ad auctions are required to be integer.} is for the situation when winning the auction and the second summation is that when losing. Similarly, the optimal action in state $(t, b, \bx)$ is
{\small
\begin{align}
\label{eq:optimal-action}
	a(t, b, \bx) &= \argmax_{0 \leq a \leq b} \Big\{
	\sum_{\delta = 0}^a m(\delta, \bx)\Big(\theta(\bx) + V(t-1, b-\delta)\Big)~+ \nonumber\\
	&~~~~~~~~~~~~~~~\sum_{\delta = a+1}^\infty m(\delta, \bx) V(t-1, b)
	\Big\},
\end{align}
}where the optimal bid action $a(t, b, \bx)$ involves three terms: $m(\delta, \bx)$, $\theta(\bx)$ and $V(t-1, \cdot)$.
$V(t,b)$ is derived by marginalizing out $\bx$:
{\small
\begin{align}
\label{eq:value-function-update}
	&V(t, b) = \int_{\bs{X}} p_x(\bx) \max_{0 \leq a \leq b} \Big\{
		\sum_{\delta = 0}^a m(\delta, \bx) \Big( \theta(\bx) + V(t-1, b-\delta) \Big) \nonumber\\
	&~~~~~~~~~~~~~~~~~~~~~~~~~~~~~~~+ \sum_{\delta = a+1}^\infty m(\delta, \bx) V(t-1, b)
		\Big\} \,d{\bx} \nonumber \\
	= & \max_{0 \leq a \leq b} \Big\{ \sum_{\delta = 0}^a \int_{\bs{X}} p_x(\bx) m(\delta, \bx) \theta(\bx) \,d{\bx} + \sum_{\delta = 0}^a V(t-1, b-\delta)~\cdot \nonumber\\
	&\int_{\bs{X}} p_x(\bx) m(\delta, \bx) \,d{\bx} + V(t-1, b) \sum_{\delta = a+1}^\infty \int_{\bs{X}} p_x(\bx) m(\delta, \bx) \,d{\bx} \Big\} \nonumber \\
	= & \max_{0 \leq a \leq b} \Big\{ \sum_{\delta = 0}^a \int_{\bs{X}} p_x(\bx) m(\delta, \bx) \theta(\bx) \,d{\bx}~+ \\
	&\sum_{\delta = 0}^a m(\delta) V(t-1, b-\delta) + V(t-1, b) \sum_{\delta = a+1}^\infty m(\delta)
	\Big\}. \nonumber
\end{align}
}


To settle the integration over $\bx$ in Eq.~(\ref{eq:value-function-update}), we consider an approximation $m(\delta, \bx) \approx m(\delta)$ by following the dependency assumption $\bx \rightarrow \theta \rightarrow a \rightarrow w \text{(winning rate)}$ in \cite{zhang2014optimal}.
Thus
\begin{align}
\label{eq:approximate-integration-over-x}
	&  \int_{\bs{X}} p_x(\bx) m(\delta, \bx) \theta(\bx) \,d{\bx} \approx m(\delta) \int_{\bs{X}} p_x(\bx) \theta(\bx) \,d{\bx} \nonumber\\
	&~~~~~~~~~~~~~~~~~~~~~~~~~~~~~~~~= m(\delta) \theta_{\text{avg}} ~,
\end{align}
where $\theta_{\text{avg}}$ is the expectation of the pCTR $\theta$, which can be easily calculated with historical data.
Taking Eq.~(\ref{eq:approximate-integration-over-x}) into Eq.~(\ref{eq:value-function-update}), we get an approximation of the optimal value function $V(t, b)$:
{\small
\begin{align}
\label{eq:value-function-final}
	V(t, b) &\approx \max_{0 \leq a \leq b} \Big\{ \sum_{\delta = 0}^a m(\delta) \theta_{\text{avg}} + \sum_{\delta = 0}^a m(\delta) V(t-1, b-\delta)~+ \nonumber\\
	&~~~~~~~~~~~~~\sum_{\delta = a+1}^\infty m(\delta) V(t-1, b)
	\Big\}.
\end{align}
}Noticing that $\sum_{\delta = 0}^\infty m(\delta, \bx) = 1$, Eq.~(\ref{eq:optimal-action}) is rewritten as
{\small
\begin{align}
\label{eq:g-function}
	&a(t, b, \bx) = \argmax_{0 \leq a \leq b} \Big\{ \sum_{\delta = 0}^a m(\delta, \bx) \Big( \theta(\bx) + V(t-1, b-\delta) \Big)- \nonumber\\
	&~~~~~~~~~~~~~~~~~~~~~~~~~~~\sum_{\delta = 0}^a m(\delta, \bx) V(t-1, b)\Big\} \nonumber \\
	&= \argmax_{0 \leq a \leq b} \Big\{ \sum_{\delta = 0}^a m(\delta, \bx) \Big(\theta(\bx) + V(t-1, b-\delta) - V(t-1, b) \Big) \Big\} \nonumber \\
	&\equiv \argmax_{0 \leq a \leq b} \Big\{ \sum_{\delta = 0}^a m(\delta, \bx) g(\delta) \Big\},
\end{align}
}where we denote $g(\delta)=\theta(\bx) + V(t-1, b-\delta) - V(t-1, b)$. From the definition, we know $V(t-1, b)$ monotonically increases w.r.t. $b$, i.e., $V(t-1, b) \geq V(t-1, b')$ where $b \geq b'$. As such, $V(t-1, b-\delta)$ monotonically decreases w.r.t. $\delta$. Thus $g(\delta)$ monotonically decreases w.r.t. $\delta$. Moreover, $g(0) = \theta(\bx) \geq 0$ and $m(\delta, \bx) \geq 0$. Here, we care about the value of $g(b)$. (i) If $g(b) \geq 0$, then $g(b') \geq g(b) \geq 0$ where $0 \leq b' \leq b$, so $m(\delta, \bx) g(\delta) \geq 0$ where $0 \leq \delta \leq b$. As a result, in this case, we have $a(t, b, \bx) = b$. (ii) If $g(b) < 0$, then there must exist an integer $A$ such that $0 \leq A < b$ and $g(A) \geq 0$, $g(A + 1) < 0$. So $m(\delta, \bx) g(\delta) \geq 0$ when $\delta \leq A$ and $m(\delta, \bx) g(\delta) < 0$ when $\delta > A$. Consequently, in this case, we have $a(t, b, \bx) = A$. In conclusion, we have
\begin{equation}
\label{eq:take-action}
\small
	a(t, b, \bx) =
	\begin{cases}
		b   &  \text{if $g(b) \geq 0$} \\
		A~\text{~~$g(A) \geq 0$ and $g(A+1) < 0$} & \text{if $g(b) < 0$}
	\end{cases}.
\end{equation}

\begin{figure}[t]
	\small
	\includegraphics[width=1.0\columnwidth]{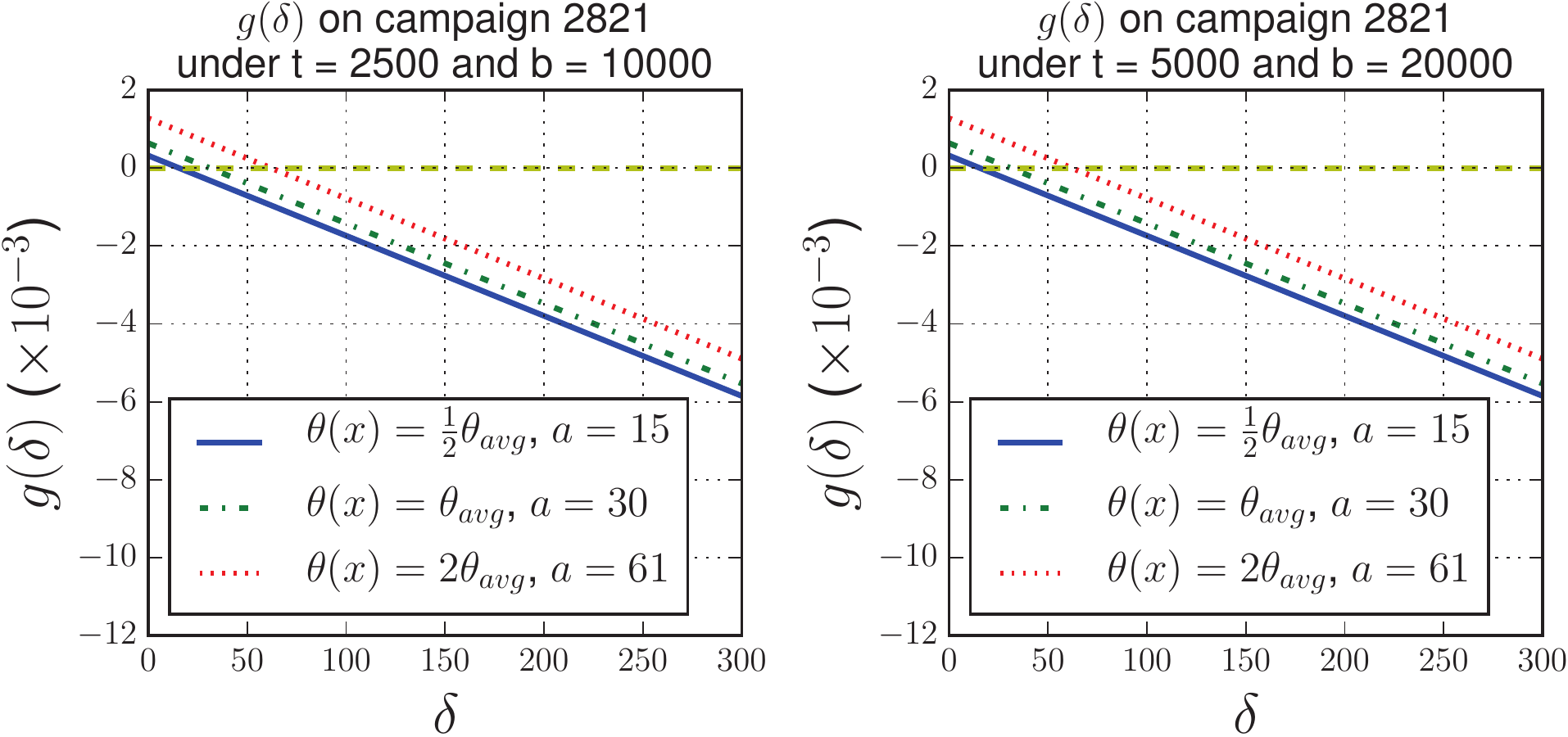}
	\vspace{-15pt}
	\caption{$g(\delta)$ on campaign 2821 as an example.}
	\label{fig:g_delta}
\end{figure}

Figure \ref{fig:g_delta} shows examples of $g(\delta)$ on campaign 2821 from iPinYou real-world dataset.
Additionally, we usually have a maximum market price $\delta_{\text{max}}$, which is also the maximum bid price. The corresponding RLB algorithm is shown in Algorithm~\ref{alg:small-scale-optimal-bidding}.

\begin{algorithm}[t]
	\caption{Reinforcement Learning to Bid}
	\label{alg:small-scale-optimal-bidding}
	{\small	
		\begin{algorithmic}[1]
			\REQUIRE
			p.d.f. of market price $m(\delta)$,
			average CTR $\theta_{\text{avg}}$,
			episode length $T$,
			budget $B$
			\ENSURE
			value function $V(t,b)$
			\STATE initialize $V(0, b)$ = 0
			\FOR{$t=1,2,\cdots, T-1$}
			\FOR{$b=0,1,\cdots, B$}
			\STATE enumerate $a_{t, b}$ from 0 to $\min(\delta_{\text{max}}, b)$ and set $V(t, b)$ via Eq. (\ref{eq:value-function-final})
			\ENDFOR
			\ENDFOR	
		\end{algorithmic}
		\begin{algorithmic}[1]
			\REQUIRE
			CTR estimator $\theta(\bx)$,
			value function $V(t, b)$,
			current state $(t_c, b_c, \bx_c)$
			\ENSURE
			optimal bid price $a_c$ in current state
			\STATE calculate the pCTR for the current bid request: $\theta_c = \theta(\bx_c)$
			\FOR{$\delta=0,1,\cdots, \min(\delta_{\text{max}}, b_c)$}
			\IF{$\theta_c + V(t_c - 1, b_c - \delta) - V(t_c - 1, b_c) \geq 0$}
			\STATE $a_c \leftarrow \delta$
			\ENDIF
			\ENDFOR
		\end{algorithmic}
	}
\end{algorithm}

\minisection{Discussion on Derived Policy}
\begin{figure}[t]
	\small
	\includegraphics[width=1.0\columnwidth]{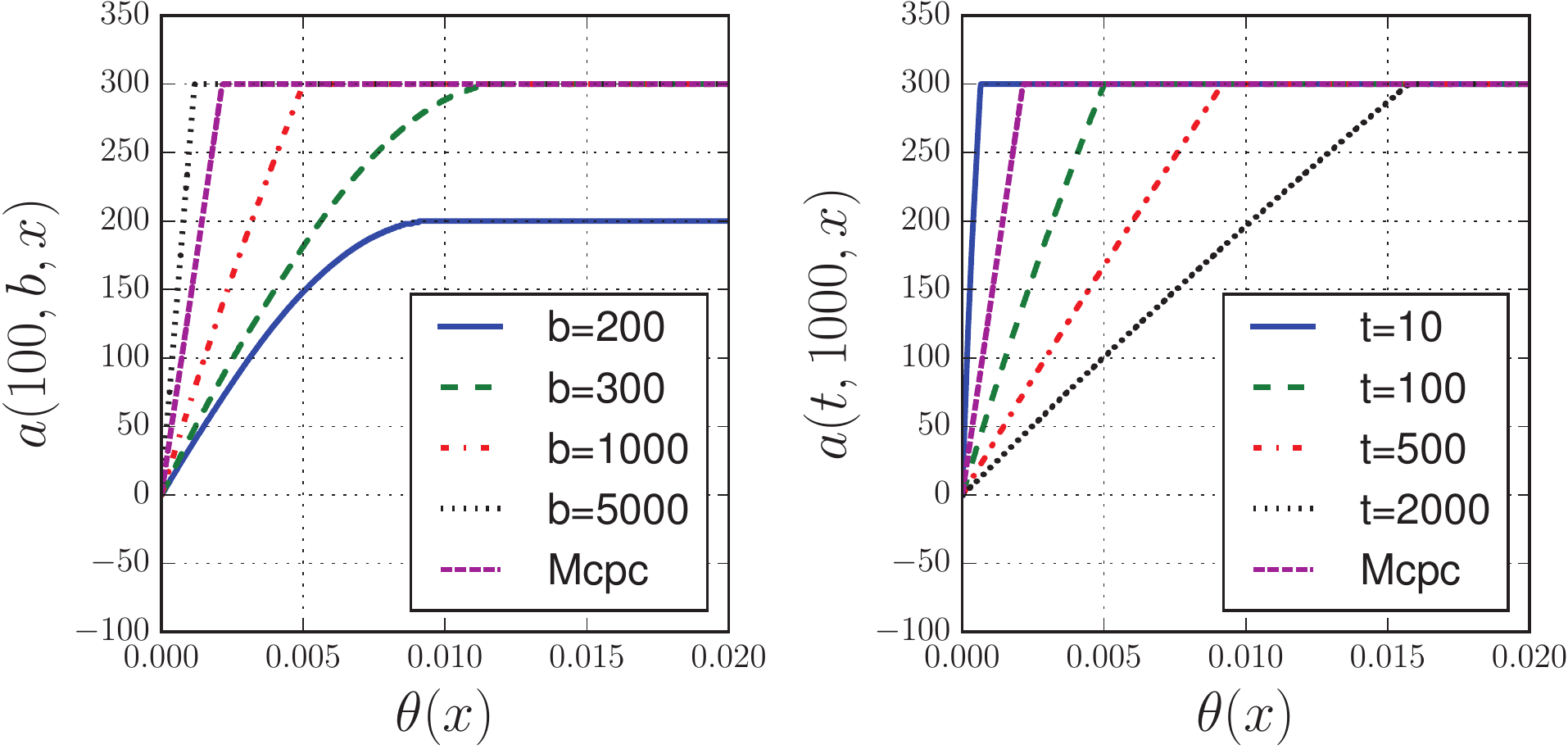}
	\vspace{-15pt}
	\caption{Example derived bidding functions.}
	\label{fig:a_theta}
\end{figure}
Contrary to the linear bidding strategies which bids linearly w.r.t. the pCTR with a static parameter \cite{perlich2012bid}, such as \tsc{Mcpc} and \tsc{Lin} discussed in Section \ref{sec:compared_meth}, our derived policy (denoted as RLB) adjusts its bidding function according to current $t$ and $b$. As shown in Figure \ref{fig:a_theta}, RLB also introduces a linear form bidding function when $b$ is large, but decreases the slope w.r.t. decreasing $b$ and increasing $t$. When $b$ is small (such as $b < 300$), RLB introduces a non-linear concave form bidding function.

\minisection{Discussion on the Approximation of $V(t, b)$}
In Eq. (\ref{eq:approximate-integration-over-x}), we take the approximation $m(\delta, \bx) \approx m(\delta)$ by following the dependency assumption $\bx \rightarrow \theta \rightarrow a \rightarrow w \text{(winning rate)}$ in \cite{zhang2014optimal} and consequently get an approximation of the optimal value function $V(t, b)$ in Eq. (\ref{eq:value-function-final}). Here, we consider a more general case where such assumption does not hold in the whole feature vector space $\bs{X}$, but holds within each individual subset. Suppose $\bs{X}$ can be explicitly divided into several segments, i.e., $\bs{X} = \sqcup_i \bs{X}_i$. The segmentation can be built by publisher, user demographics etc. For each segment $\bs{X}_i$, we take the approximation $m(\delta, \bx) \approx m_i(\delta)$ where $\bx \in \bs{X}_i$. As such, we have
{\small
\begin{align}
	&\int_{\bs{X}} p_x(\bx) m(\delta, \bx) \theta(\bx) \,d{\bx}
	= \sum_{i} \int_{\bs{X}_i} p_x(\bx) m(\delta, \bx) \theta(\bx) \,d{\bx}  \nonumber \\
	\approx& \sum_{i} m_i(\delta) \int_{\bs{X}_i} p_x(\bx) \theta(\bx) \,d{\bx}
	= \sum_{i} m_i(\delta) (\theta_{\text{avg}})_i P(\bx \in \bs{X}_i).  \nonumber
\end{align}
}

\subsection{Handling Large-Scale Issues}\label{sec:large-scale-models}
Algorithm \ref{alg:small-scale-optimal-bidding} gives a solution to the optimal policy. However, when it comes to the real-world scale, we should also consider the complexity of the algorithm. Algorithm \ref{alg:small-scale-optimal-bidding} consists of two stages. The first one is about updating the value function $V(t, b)$, while the second stage is about taking the optimal action for current state based on $V(t, b)$. We can see that the main complexity is on the first stage. Thus we focus on the first stage in this section.
Two nested loops in the first stage lead the time complexity to $O(TB)$. As for the space complexity, we need to use a two-dimensional table to store $V(t, b)$, which will later be used when taking action. Thus the space complexity is $O(TB)$.

In consideration of the space complexity and the time complexity, Algorithm \ref{alg:small-scale-optimal-bidding} can only be applied to small-scale situations. When we confront the situation where $T$ and $B$ are very large, which is a common case in real world, there will probably be not enough resource to get the exact value of $V(t, b)$ for every $(t, b) \in \{0,\cdots,T\} \times \{0,\cdots,B\}$.

With restricted computational resources, one may not be able to go through the whole value function update. Thus we propose some parameterized models to fit the value function on small data scale, i.e., $\{0,\cdots,T_0\} \times \{0,\cdots,B_0\}$, and generalize to the large data scale $\{0,\cdots,T\} \times \{0,\cdots,B\}$.


Good parameterized models are supposed to have low deviation to the exact value of $V(t, b)$ for every $(t, b) \in \{0,\cdots,T\} \times \{0,\cdots,B\}$. That means low root mean square error (RMSE) in the training data and good generalization ability.



Basically, we expect the prediction error of $\theta(\bx) + V(t-1, b-\delta) - V(t-1, b)$ from Eq.~(\ref{eq:g-function}) in the training data to be low in comparison to the average CTR $\theta_{\text{avg}}$.
For most $(t, b)$, $V(t, b)$ is much larger than $\theta_{\text{avg}}$. For example, if the budget $b$ is large enough, $V(t, b)$ is with the same scale of $t \times \theta_{\text{avg}}$. Therefore, if we take $V(t, b)$ as our target to approximate, it is difficult to give a low deviation in comparison to $\theta_{\text{avg}}$. Actually, when calculating $a(t, b, \bx)$ in Eq.~(\ref{eq:g-function}), we care about the value of $V(t-1, b-\delta) - V(t-1, b)$ rather than $V(t-1, b-\delta)$ or $V(t-1, b)$. Thus here we introduce a new function of value differential $D(t, b) = V(t, b + 1) - V(t, b)$ to replace the role of $V(t, b)$ by
\begin{equation}\small
V(t-1, b-\delta) - V(t-1, b) = -\sum_{\delta'=1}^{\delta} D(t-1, b-\delta').
\end{equation}
\begin{figure}[t]
	\includegraphics[width=1.0\columnwidth]{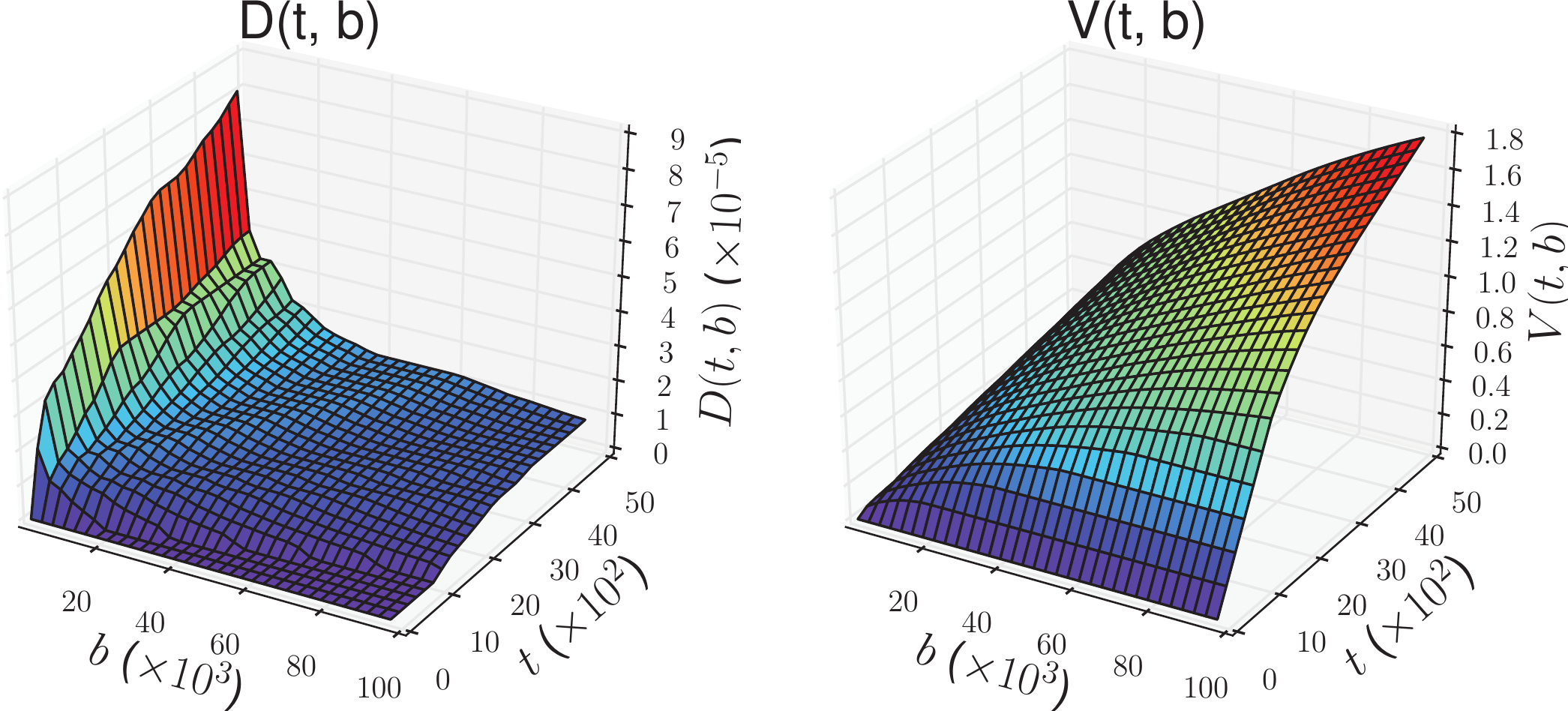}
    \caption{$D(t, b)$ and $V(t, b)$ on campaign 3427.}
	\label{fig:DV_3d}
\end{figure}

\begin{figure}[t]
	\includegraphics[width=1.0\columnwidth]{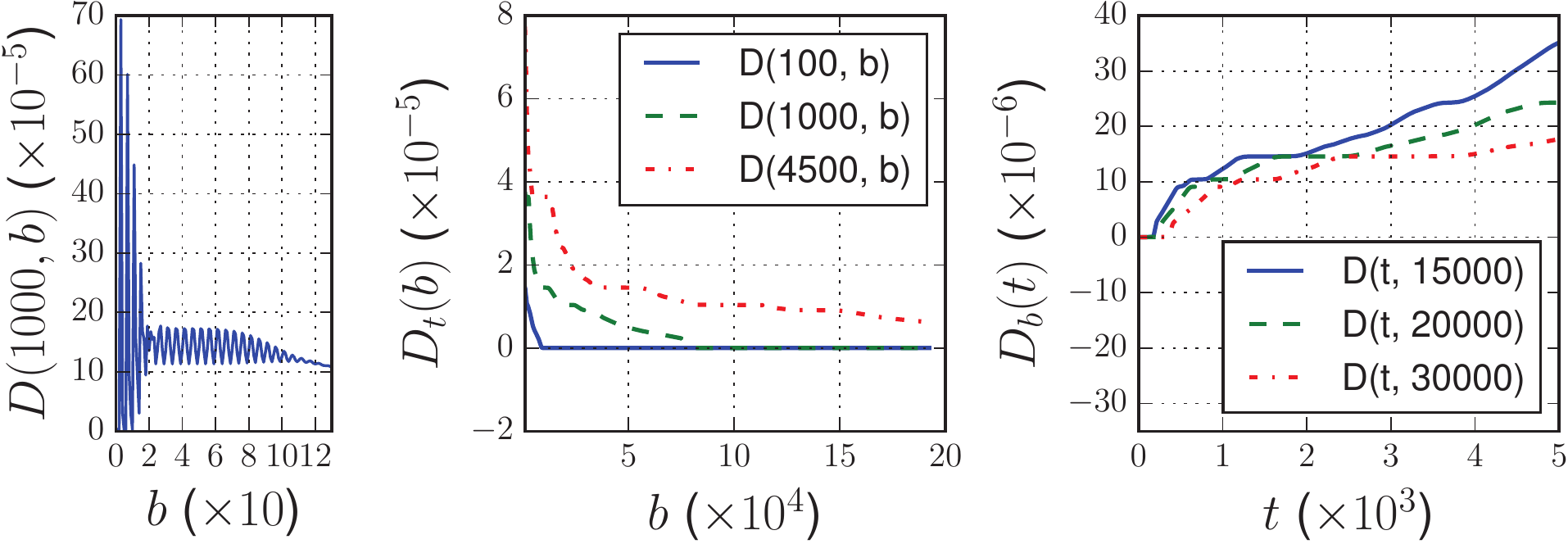}
	\caption{Analysis of $D(t, b)$ on campaign 3386.}
	\label{fig:D_t_b_analysis}
\end{figure}

Figure~\ref{fig:DV_3d} illustrates the value of $D(t, b)$ and $V(t, b)$ on the data of an example campaign.
In Figure \ref{fig:D_t_b_analysis}, we use the campaign 3386 from iPinYou real-world dataset as an example and show some interesting observations of $D(t, b)$ (other campaigns are similar). At first, for a given $t$, we consider $D(t, b)$ as a function of $b$ and denote it as $D_t(b)$. $D_t(b)$ fluctuates heavily when $b$ is very small, and later keeps decreasing to 0. Similarly, for a given $b$, we have $D_b(t)$ as a function of $t$ and it keeps increasing. Moreover, both $D_t(b)$ and $D_b(t)$ are obviously nonlinear. Consequently, we apply the neural networks to approximate them for large-scale $b$ and $t$.

As a widely used solution \cite{boyan1995generalization,sutton1998introduction}, here we take a fully connected neural network with several hidden layers as a non-linear approximator. The input layer has two nodes for $t$ and $b$.
The output layer has one node for $D(t, b)$ without activation function. As such, the neural network corresponds to a non-linear function of $t$ and $b$, denoted as $\text{NN}(t, b)$.


\minisection{Coarse-to-fine Episode Segmentation Model} Since the neural networks do not guarantee good generalization ability and may suffer from overfitting, and also to avoid directly modeling $D(t, b)$ or $V(t, b)$, we explore the feasibility of mapping unseen states ($t>T_0$ and $b>B_0$) to acquainted states ($t \leq T_0$ and $b \leq B_0$) rather than giving a global parameterized representation.
Similar to budget pacing, we have the first simple implicit mapping method where we can divide the large episode into several small episodes with length $T_0$ and within each large episode we allocate the remaining budget to the remaining small episodes. 
If the agent does not spend the budget allocated for the small episode, it will have more allocated money for the rest of the small episodes in the large episode.

\begin{figure}[t]
	\small
	\centering
	\includegraphics[width=.9\columnwidth]{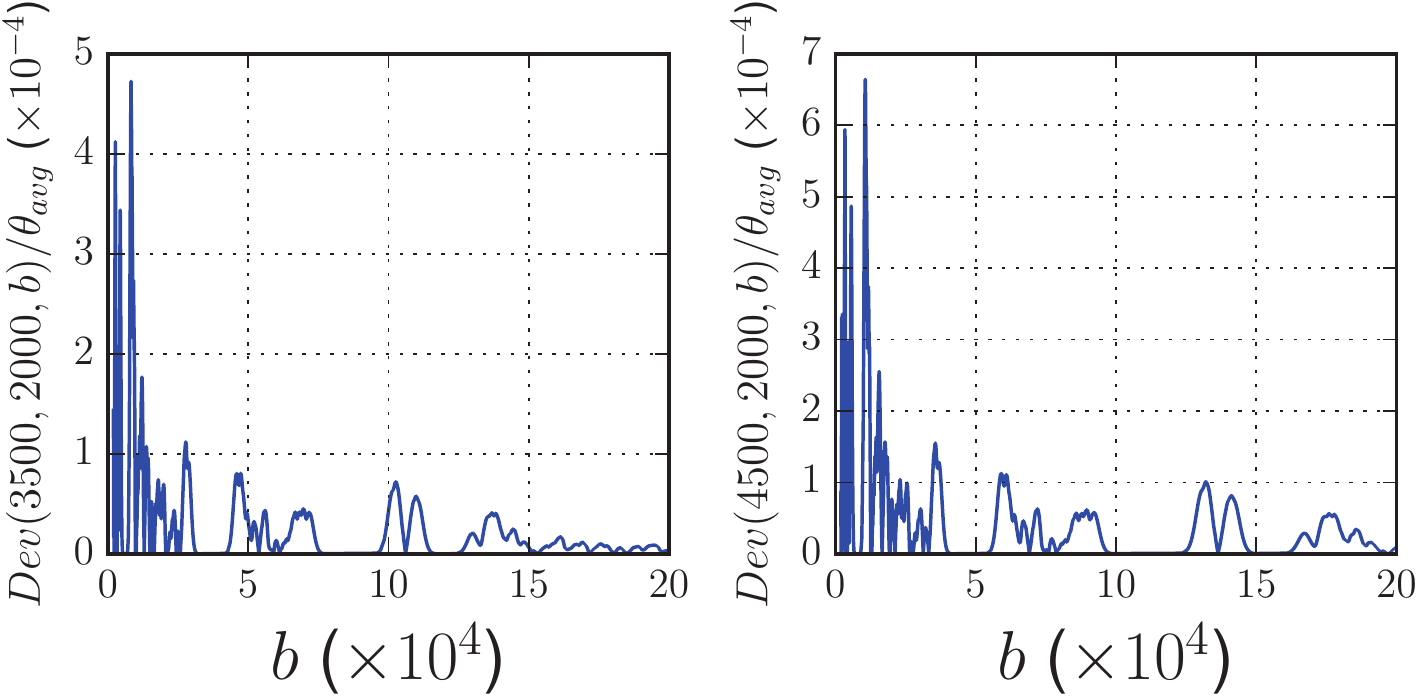}
	\caption{$Dev(t, T_0, b)$ on campaign 3427 as an example.}
	\label{fig:Dev}
\end{figure}

\minisection{State Mapping Models} Also we consider explicit mapping methods. At first, because $D_t(b)$ keeps decreasing and $D_b(t)$ keeps increasing, then for $D(t, b)$ where $t$ and $b$ are large, there should be some points $\{(t', b')\}$ where $t' \leq T_0$ and $b' \leq B_0$ such that $D(t', b') = D(t, b)$ as is shown in Figure \ref{fig:DV_3d}, which confirms the existence of the mapping for $D(t, b)$. Similarly, $a(t, b, \bx)$ decreases w.r.t. $t$ and increases w.r.t. $b$, which can be seen in Figure \ref{fig:a_theta} and is consistent with intuitions. Thus the mapping for $a(t, b, \bx)$ also exists. From the view of practical bidding, when the remaining number of auctions are large and the budget situation is similar, given the same bid request, the agent should give a similar bid price (see Figure~\ref{fig:g_delta}). We consider a simple case that $b/t$ represents the budget condition. Then here we have two linear mapping forms: (i) map $a(t, b, \bx)$ where $t > T_0$ to $a(T_0, \frac{b}{t} \times T_0, \bx)$. (ii) map $D(t, b)$ where $t > T_0$ to $D(T_0, \frac{b}{t} \times T_0)$. Denote $|D(t, b) - D(T_0, \frac{b}{t} \times T_0)|$ as $Dev(t, T_0, b)$. Figure \ref{fig:Dev} shows that the deviations of the simple linear mapping method are low enough ($<10^{-3}\theta_{\text{avg}}$).


\section{Experimental Setup}\label{sec:exp-setup}
\subsection{Datasets}
Two real-world datasets are used in our experimental study, namely iPinYou and YOYI.
\begin{description}
\item{\textbf{iPinYou}} is one of the mainstream RTB ad companies in China.
The whole dataset comprises 19.5M impressions, 14.79K clicks and 16.0K \text{CNY} expense on 9 different campaigns over 10 days in 2013. We follow \cite{zhang2015statistical} for splitting the train/test sets and feature engineering.
\item{\textbf{YOYI}}
is a leading RTB company focusing on multi-device display advertising in China.
YOYI dataset comprises 441.7M impressions, 416.9K clicks and 319.5K CNY expense during 8 days in Jan.~2016. The first 7 days are set as the training data while the last day is set as the test data.
\end{description}


For experiment reproducibility we publicize our code\footnote{The experiment code is available at \url{https://github.com/han-cai/rlb-dp} and iPinYou dataset is available at \url{http://data.computational-advertising.org}.}. In the paper we mainly report results on iPinYou dataset, and further verify our algorithms over the YOYI dataset as supplementary.



%
\vspace{10pt}
\subsection{Evaluation Methods}
The evaluation is from the perspective of an advertiser's campaign with a predefined budget and lifetime (episode length).

\minisection{Evaluation metrics}
The main goal of the bidding agent is to optimise the campaign's KPI (e.g., clicks, conversions, revenue, etc.) given the campaign budget. In our work, we consider the number of acquired clicks as the KPI, which is set as the primary evaluation measure in our experiments. We also analyze other statistics such as win rate, cost per mille impressions (CPM) and effective cost per click (eCPC).

\minisection{Evaluation flow}
We mostly follow \cite{zhang2014optimal} when building the evaluation flow, except that we divide the test data into episodes. Specifically,
the test data is a list of records, each of which consists of the bid request feature vector, the market price and the user response (click) label. We divide the test data into episodes, each of which contains $T$ records and is allocated with a budget $B$. Given the CTR estimator and the bid landscape forecasting, the bidding strategy goes through the test data episode by episode. Specifically, the bidding strategy generates a price for each bid request (the bid price cannot exceed current budget). If the bid price is higher than or equal to the market price of the bid request, the advertiser wins the auction and then receives the market price as cost and the user click as reward and then updates the remaining auction number and budget.

\minisection{Budget constraints}
Obviously, if the allocated budget $B$ is too high, the bidding strategy can simply give a very high bid price each time to win all clicks in the test data. Therefore, in evaluation, budget constraints should not be higher than the historic total cost of the test data. We determine the budget $B$ in this way: $B = \text{CPM}_{\text{train}} \times 10^{-3} \times T \times c_0$, where $\text{CPM}_{\text{train}}$ is the cost per mille impressions in the training data and $c_0$ acts as the budget constraints parameter. Following previous work \cite{zhang2014optimal,zhang2015statistical}, we run the evaluation with $c_0 = 1/32, 1/16, 1/8, 1/4, 1/2$.

\minisection{Episode length}
The episode auction number $T$ influences the complexity of our algorithms. When $T$ is high, the original Algorithm~\ref{alg:small-scale-optimal-bidding} is not capable of working with limited resources, which further leads to our large-scale algorithms. For the large-scale evaluation, we set $T$ as 100,000, which corresponds to a real-world 10-minute auction volume of a medium-scale RTB ad campaign. And for the small-scale evaluation, we set the episode length as 1,000. In addition, we run a set of evaluations with $c_0 = 0.2$ and the episode length $T = 200, 400, 600, 800, 1000$ to give a more comprehensive performance analysis.

\subsection{Compared Methods}
\label{sec:compared_meth}
The following bidding policies are compared with the same CTR estimation component which is a logistic regression model and the same bid landscape forecasting component which is a non-parametric method, as described in Section \ref{sec:related}:
\begin{description}
\item[SS-MDP] is based on \cite{amin2012budget}, considering the bid landscape but ignoring the feature vector of bid request when giving the bid price. Although we regard this model as the state-of-the-art, it is proposed to work on keyword-level bidding in sponsored search, which makes it not fine-grained enough to compare with RTB display advertising strategies.
\item[Mcpc] gives its bidding strategy as $a_{\tsc{Mcpc}}(t, b, \bx) =$ CPC $\times$ $\theta(\bx)$, which matches some advertisers' requirement of maximum CPC (cost per click).
\item[Lin] is a linear bidding strategy w.r.t. the pCTR:  $a_{\tsc{Lin}}(t, b, \bx)$ $= b_0 \frac{\theta(\bx)}{\theta_{\text{avg}}}$, where $b_0$ is the basic bid price and is tuned using the training data \cite{perlich2012bid}. This is the most widely used model in industry.
\item[RLB] is our proposed model for the small-scale problem as shown in Algorithm~\ref{alg:small-scale-optimal-bidding}.
\item[RLB-NN] is our proposed model for the large-scale problem, which uses the neural network $\text{NN}(t, b)$ to approximate $D(t, b)$.
\item[RLB-NN-Seg] combines the neural network with episode segmentation. For each small episode, the allocated budget is $B_s = B_r / N_r$ where $B_r$ is the remaining budget of the current large episode and $N_r$ is the remaining number of small episodes in the current large episode. Then RLB-NN is run for the small episode. It corresponds to the coarse-to-fine episode segmentation model discussed in Section~\ref{sec:large-scale-models}.
\item[RLB-NN-MapD] combines the neural network with the mapping of $D(t, b)$. That is: (i) $D(t, b) = \text{NN}(t, b)$ where $t \leq T_0$. (ii) $D(t, b) = \text{NN}(T_0, \frac{b}{t} \times T_0)$ where $t > T_0$.
\item[RLB-NN-MapA] combines the neural network with the mapping of $a(t, b, \bx)$. That is: $a(t, b, \bx) = a(T_0, \frac{b}{t} \times T_0, \bx)$ where $t > T_0$. The last two models correspond to the state mapping models discussed in Section~\ref{sec:large-scale-models}.
\end{description}

\begin{figure}[t]
	\small
	\includegraphics[width=1.0\columnwidth]{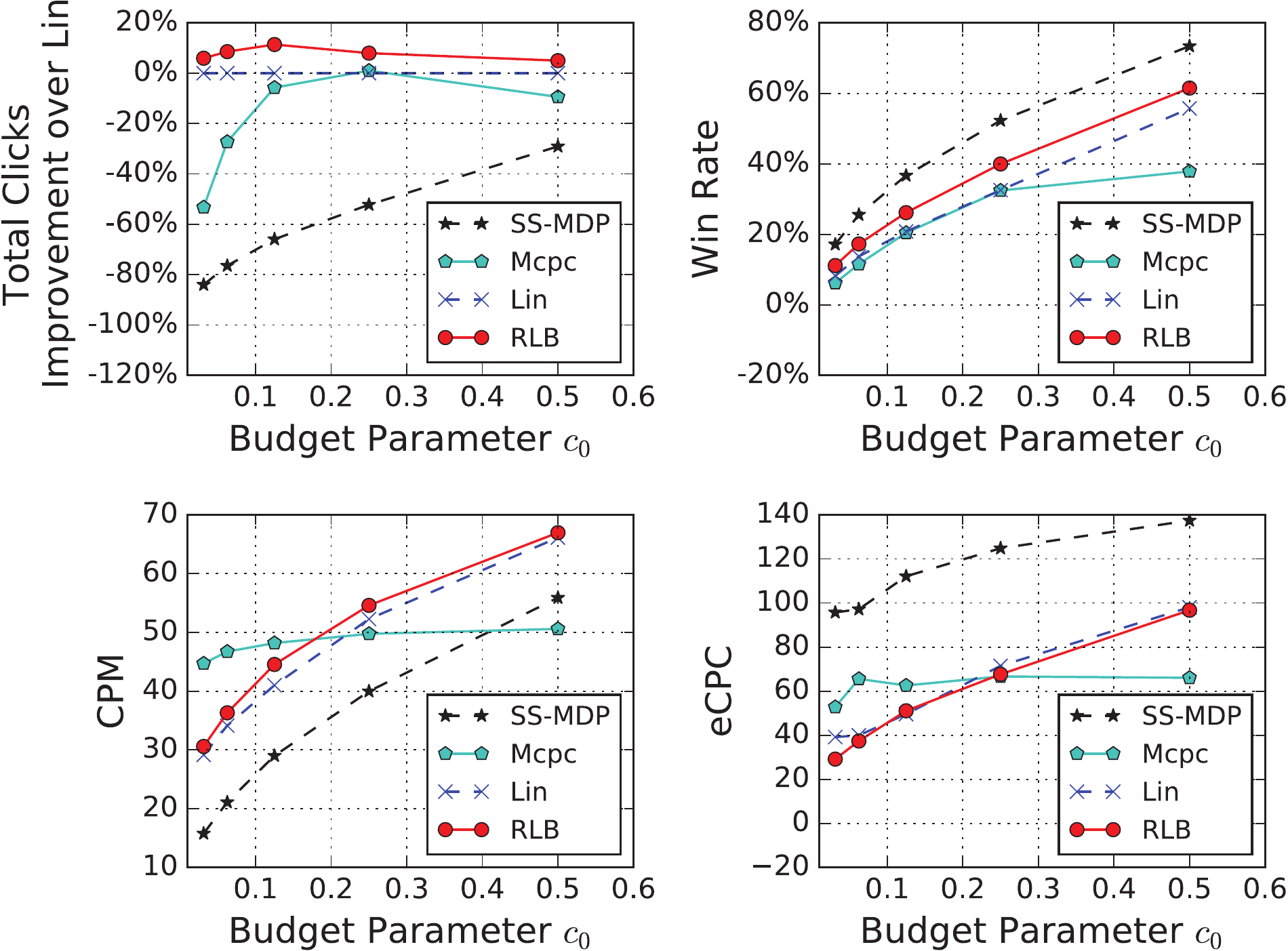}
    \caption{Overall performance on iPinYou under $T = 10^3$ and different budget conditions.}
	\label{fig:s_perf_overall_ipinyou}
\end{figure}

\begin{table}[t]
\small
	\centering
	\caption{Click improvement of \tsc{RLB} over \tsc{Lin} for each campaign under $T = 10^3$ and different budget conditions.}\label{tab:RLB_vs_Lin}
	\vspace{15pt}
	\begin{tabular}{c | rrrrr}
		iPinYou & 1/32 & 1/16 & 1/8 & 1/4 & 1/2\\
		\hline
		1458 & 4.66\% & 3.96\% & 3.25\% & 0.21\% & 1.02\% 		\\
		2259 & 114.29\% & 35.29\% & 9.09\% & 32.56\% & 22.22\% 	\\
		2261 & 25.00\% & 6.25\% & -3.70\% & 6.82\% & 0.00\% 	\\
		2821 & 20.00\% & 11.86\% & 27.27\% & 29.36\% & 12.97\% 	\\
		2997 & 23.81\% & 54.55\% & 85.26\% & 13.04\% & 3.18\%	\\
		3358 & 2.42\% & 3.30\% & 0.87\% & 3.02\% & 0.40\%		\\
		3386 & 8.47\% & 22.47\% & 13.24\% & 14.57\% & 13.40\%	\\
		3427 & 7.58\% & 10.04\% & 12.28\% & 6.88\% & 5.34\%		\\
		3476 & -4.68\% & -3.79\% & 2.50\% & 5.43\% & 0.72\%		\\
		\hline
		Average & 22.39\% & 15.99\% & 16.67\% & 12.43\% & 6.58\%\\
		\hline
		\hline
		YOYI & 3.89\% & 2.26\% & 7.41\% & 3.48\% & 1.71\%		\\
		\hline
	\end{tabular}
\end{table}

\section{Experimental Results}\label{sec:exp-res}
In this section we present the experimental results on small- and large-scale data settings respectively.
\subsection{Small-Scale Evaluation}


The performance comparison on iPinYou dataset under $T = 1000$ and different budget conditions are reported in Figure \ref{fig:s_perf_overall_ipinyou}. In the comparison on total clicks (upper left plot), we find that (i) our proposed model \tsc{RLB} performs the best under every budget condition, verifying the effectiveness of the derived algorithm for optimizing attained clicks. (ii) \tsc{Lin} has the second best performance, which is a widely used bidding strategy in industry \cite{perlich2012bid}. (iii) Compared to \tsc{RLB} and \tsc{Lin}, \tsc{Mcpc} does not adjust its strategy when the budget condition changes. Thus it performs quite well when $c_0 \geq 1/4$ but performs poorly on very limited budget conditions, which is consistent with the discussion in Section~\ref{sec:related}. (iv) \tsc{SS-MDP} gives the worst performance, since it is unaware of the feature information of each bid request, which shows the advantages of RTB display advertising.

As for the comparison on win rate, CPM and eCPC, we observe that (i) under every budget condition, \tsc{SS-MDP} keeps the highest win rate. The reason is that \tsc{SS-MDP} considers each bid request equally, thus its optimization target is equivalent to the number of impressions. Therefore, its win rate should be the highest. (ii) \tsc{Lin} and \tsc{RLB} are very close in comparison on CPM and eCPC. \tsc{RLB} can generate a higher number of clicks with comparable CPM and eCPC against \tsc{Lin} because \tsc{RLB} effectively spends the budget according to the market situation, which is unaware of by \tsc{Lin}.



\begin{figure}[t]
	\small
	\includegraphics[width=1.0\columnwidth]{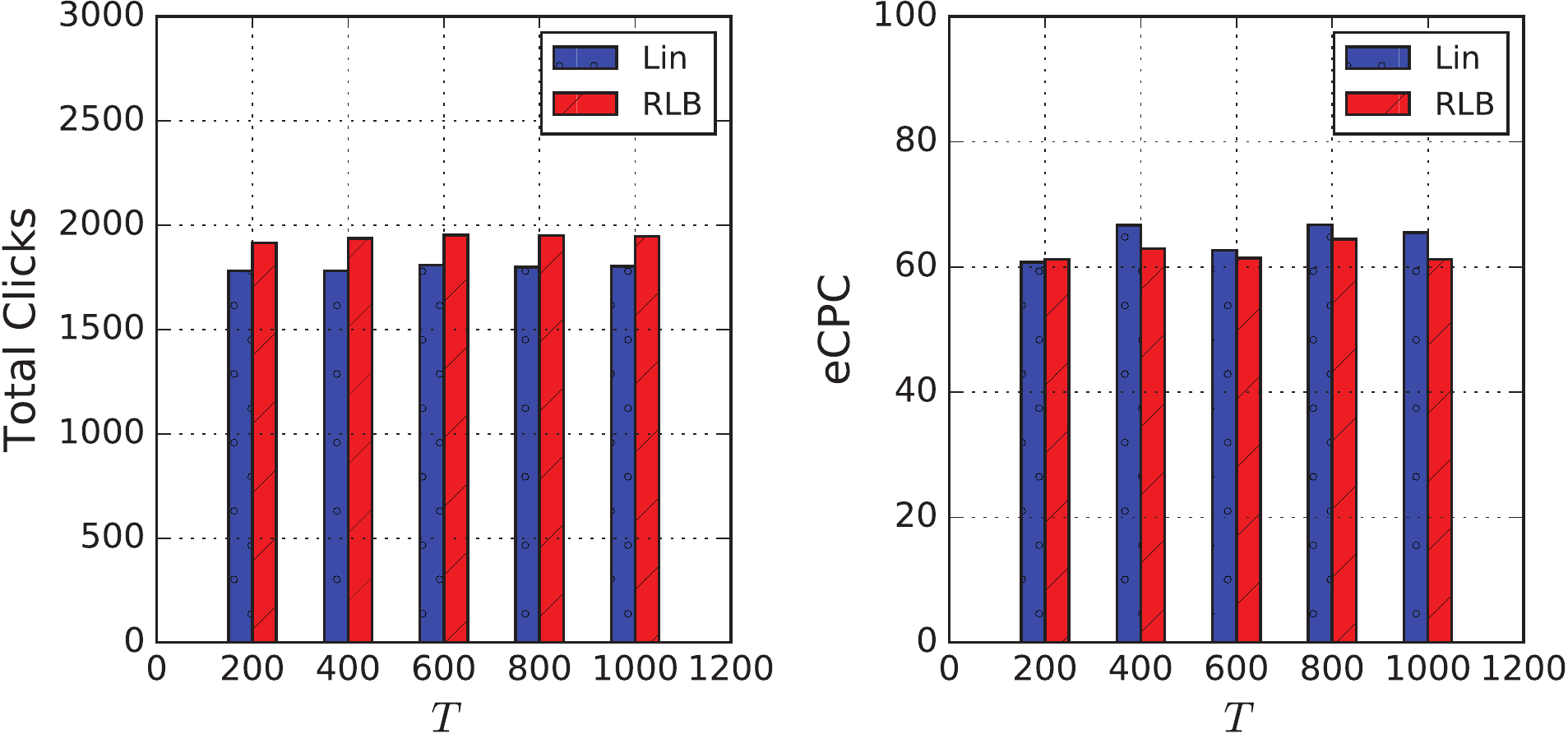}
    \caption{Overall performance comparison on iPinYou under $c_0 = 0.2$ and different $T$'s.}
	\label{fig:s_perf_Ts_ipinyou}
\end{figure}

Table \ref{tab:RLB_vs_Lin} provides a detailed performance on clicks of \tsc{RLB} over \tsc{Lin} under various campaigns and budget conditions. Among all 50 settings, \tsc{RLB} wins \tsc{Lin} in 46 (92\%), ties in 1 (2\%) and loses in 3 (6\%) settings. It shows that \tsc{RLB} is robust and significantly outperforms \tsc{Lin} in the vast majority of the cases. Specifically, for 1/8 budget, \tsc{RLB} outperforms \tsc{Lin} by 16.7\% on iPinYou data and 7.4\% on YOYI data. Moreover, Figure \ref{fig:s_perf_Ts_ipinyou} shows the performance comparison under the same budget condition ($c_0 = 0.2$) and different episode lengths. The findings are similar to the above results. Compared to \tsc{Lin}, \tsc{RLB} can attain more clicks with similar eCPC. Note that, in offline evaluations the total auction number is stationary, larger episode length also means smaller episode number. Thus the total click numbers in Figure~\ref{fig:s_perf_Ts_ipinyou} do not increase largely w.r.t. $T$.

\tsc{SS-MDP} is the only model that ignores the feature information of the bid request, thus providing a poor overall performance. Table~\ref{tab:all_compare_table} reports in detail the clicks along with the AUC of the CTR estimator for each campaign. We find that when the performance of the CTR estimator is relatively low (AUC $< 70\%$), e.g., campaign 2259, 2261, 2821, 2997, the performance of \tsc{SS-MDP} on clicks is quite good in comparison to \tsc{Mcpc} and \tsc{Lin}. By contrast, when the performance of the CTR estimator gets better, other methods which utilize the CTR estimator can attain much more clicks than \tsc{SS-MDP}.

\begin{table}[t]
\small
	\centering
	\caption{Detailed AUC and clicks ($T = 10^3$ and $c_0 = 1/16$).}\label{tab:all_compare_table}
	\begin{tabular}{c | c | cccc}
		iPinYou & AUC of $\theta(\bx)$ & \tsc{SS-MDP} & \tsc{Mcpc} & \tsc{Lin} & \tsc{RLB}\\
		\hline
		1458 & 97.73\% & 42 & 405 & 455 & 473 		\\
		2259 & 67.90\% & 13 & 11 & 17 & 23 		\\
		2261 & 62.16\% & 16 & 12 & 16 & 17 		\\
		2821 & 62.95\% & 49 & 38 & 59 & 66 		\\
		2997 & 60.44\% & 116 & 82 & 77 & 119		\\
		3358 & 97.58\% & 15 & 144 & 212 & 219		\\
		3386 & 77.96\% & 24 & 56 & 89 & 109			\\
		3427 & 97.41\% & 20 & 178 & 279 & 307		\\
		3476 & 95.84\% & 38 & 103 & 211 & 203		\\
		\hline
		Average & 80.00\% & 37 & 114 & 157 & 170	\\
		\hline
		\hline
		YOYI & 87.79\% & 120 & 196 & 265 & 271		\\
		\hline
	\end{tabular}
\end{table}
\subsection{Large-Scale Evaluation}
In this section, we first run the value function update in Algorithm~\ref{alg:small-scale-optimal-bidding} under $T_0 = 10,000$ and $B_0 = \text{CPM}_{\text{train}} \times 10^{-3} \times T_0 \times 1/2$, then train a neural network with the attained data $(t, b, D(t, b))$ (where $(t, b) \in \{0,\cdots,T_0\} \times \{0,\cdots,B_0\}$).
Here we use a fully connected neural network with two hidden layers which use tanh activation function. The first hidden layer has 30 hidden nodes and the second one has 15 hidden nodes.
Next, we apply the neural network to run bidding under $T = 100,000$ and $B = \text{CPM}_{\text{train}} \times 10^{-3} \times T \times c_0$. In addition, \tsc{SS-MDP} is not tested in this experiment because it suffers from scalability issues and will have a similarly low performance as in the small-scale evaluation.



Table~\ref{tab:rmse_nn} shows the performance of the neural network on iPinYou and YOYI. We can see that the RMSE is relatively low in comparison to $\theta_{\text{avg}}$, which means that the neural network can provide a good approximation to the exact algorithm when the agent comes to a state $(t, b, \bx)$ where $(t, b) \in \{0,\cdots,T_0\} \times \{0,\cdots,B_0\}$.

\begin{figure}[t]
	\small
	\includegraphics[width=1.0\columnwidth]{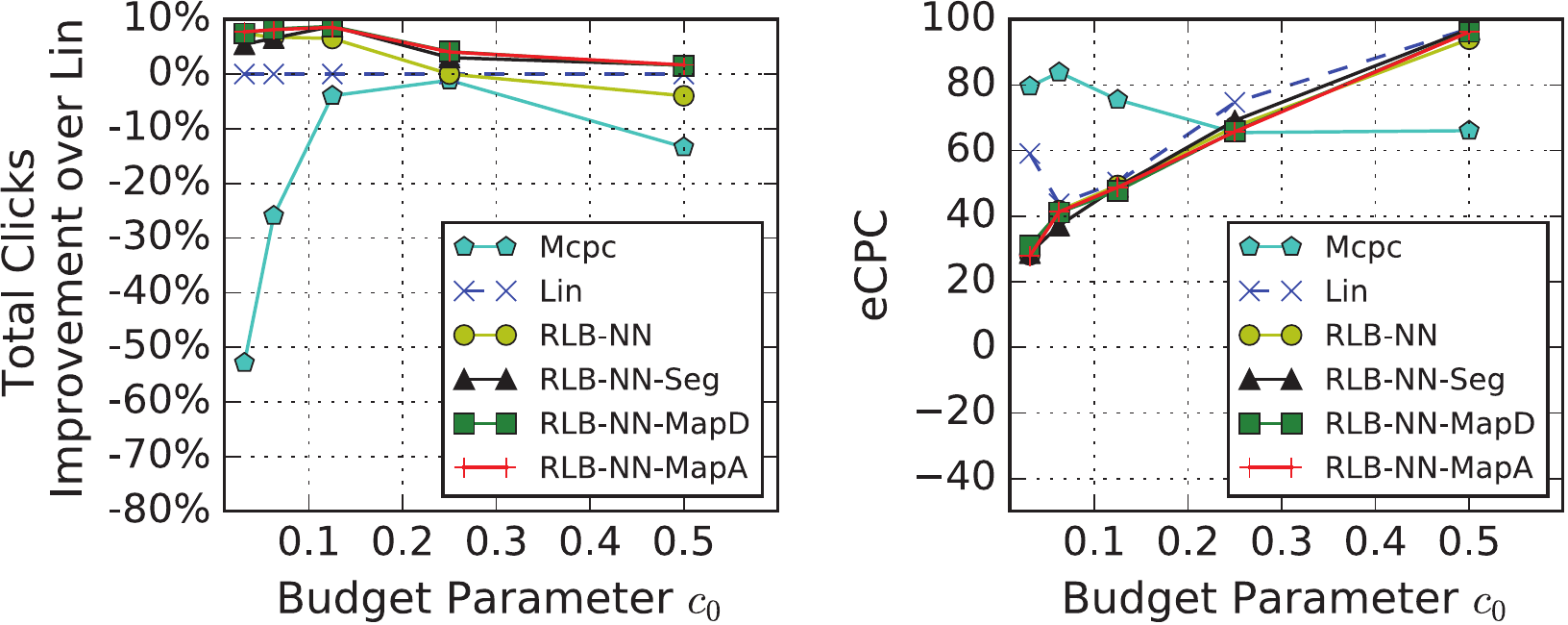}
    \caption{Overall performance on iPinYou under $T = 10^5$ and different budget conditions.}
	\label{fig:l_perf_overall_ipinyou}
\end{figure}

Figure \ref{fig:l_perf_overall_ipinyou} shows the performance comparison on iPinYou under $T = 100,000$ and different budget conditions. We observe that (i)
\tsc{Mcpc} has a similar performance to that observed in small-scale situations. (ii) For total clicks, \tsc{RLB-NN} performs better than \tsc{Lin} under $c_0=$ 1/32, 1/16, 1/8 and performs worse than \tsc{Lin} under $c_0=$ 1/2, which shows that the generalization ability of the neural network is satisfactory only in small scales. For relatively large scales, the generalization of \tsc{RLB-NN} is not reliable. (iii) Compared to \tsc{RLB-NN}, the 3 sophisticated algorithms \tsc{RLB-NN-Seg}, \tsc{RLB-NN-MapD} and \tsc{RLB-NN-MapA} are more robust and outperform \tsc{Lin} under every budget condition. They do not rely on the generalization ability of the approximation model, therefore their performance is more stable. The results clearly demonstrate that they are effective solutions for the large-scale problem. (iv) As for eCPC, all models except from \tsc{Mcpc} are very close, thus making the proposed \tsc{RLB} algorithms practically effective.



\begin{table}[t]
\small
	\centering
	\caption{Approximation performance of the neural network.}\label{tab:rmse_nn}
    \begin{tabular}{c | c | c }
		 & iPinYou & YOYI\\
		\hline
		RMSE ($\times 10^{-6}$) & 0.998 & 1.263\\
		RMSE / $\theta_{\text{avg}}$ ($\times 10^{-4}$) & 9.404 & 11.954\\
		\hline
	\end{tabular}
\end{table}

\section{Online Deployment and A/B Test}
Our proposed RLB model is deployed and tested in a live environment provided by Vlion DSP. The deployment environment is based on HP ProLiant DL360p Gen8 servers. A 5-node cluster is utilized for the bidding agent, where each node is in CentOS release 6.3, with 6 core Intel Xeon CPU E5-2620 (2.10GHz) and 64GB RAM. The model is implemented in Lua with Nginx.

\begin{figure}[t]
	\small
	\includegraphics[width=1.0\columnwidth]{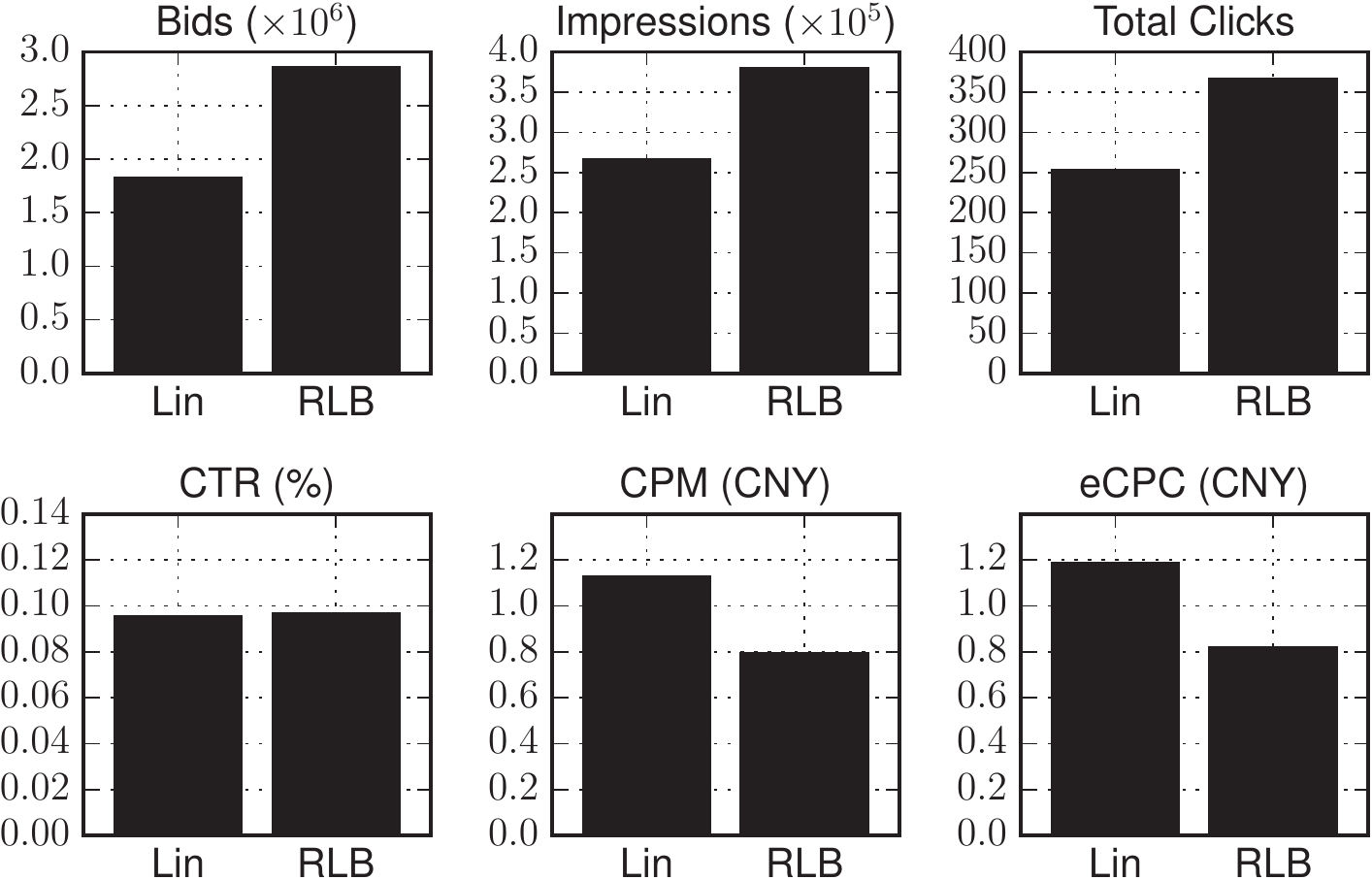}
	\caption{Online A/B testing results.}
	\label{fig:online_overall}
\end{figure}

\begin{figure}[t]
	\small
	\includegraphics[width=1.0\columnwidth]{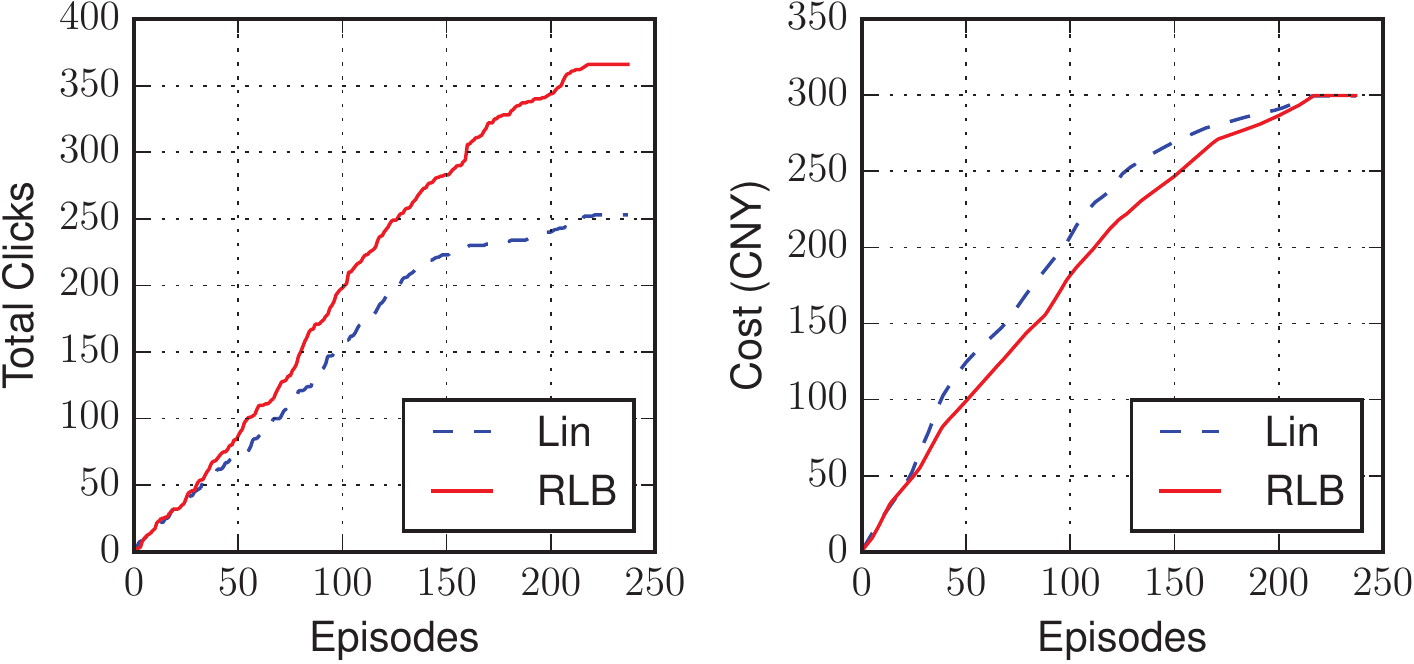}
	\caption{Total clicks and cost increase over episodes.}
	\label{fig:online_time}
\end{figure}

The compared bidding strategy is \tsc{Lin} as discussed in Section \ref{sec:compared_meth}. The optimization target is click. The two compared methods are given the same budget, which is further allocated to episodes. Unlike offline evaluations, the online evaluation flow stops only when the budget is exhausted. Within an episode, a maximum bid number $T$ is set for each strategy to prevent overspending too much. Specifically, $T$ is mostly determined by the allocated budget for the episode $B$, previous CPM and win rate: $T = B / \text{CPM} / \text{win rate} \times 10^3$. The possible available auction number during the episode is also considered when determining $T$. 
The agent keeps the remaining bid number and budget, which we consider as $t$ and $b$ respectively. Note that the remaining budget may have some deviation due to latency. The latency is typically less than 100ms, which is negligible. We test over 5 campaigns during 25-28th of July, 2016. All the methods share the same previous 7-day training data, and the same CTR estimator which is a logistic regression model trained with FTRL.
The bid requests of each user are randomly sent to either method. The overall results are presented in Figure \ref{fig:online_overall}, while the click and cost performances w.r.t. time are shown in Figure \ref{fig:online_time}.

From the comparison, we observe the following:
(i) with the same cost, \tsc{RLB} achieves lower eCPC than \tsc{Lin}, and thus more total clicks, which shows the cost effectiveness of \tsc{RLB}.
(ii) \tsc{RLB} provides better planning than \tsc{Lin}: the acquired clicks and spent budget increase evenly across the time.
(iii) With better planning, \tsc{RLB} obtains lower CPM than \tsc{Lin}, yielding more bids and more winning impressions.
(iv) With lower CPM on cheap cases, \tsc{RLB} achieves a close CTR compared to \tsc{Lin}, which leads to superior performance.
In summary, the online evaluation demonstrates the effectiveness of our proposed RLB model for optimizing attained clicks with a good pacing.

\section{Conclusions}\label{sec:conclusion}
In this paper, we proposed a model-based reinforcement learning model (RLB) for learning the bidding strategy in RTB display advertising. The bidding strategy is naturally defined as the policy of making a bidding action given the state of the campaign's parameters and the input bid request information. With an MDP formulation, the state transition and reward function are captured via modeling the auction competition and user click, respectively. The optimal bidding policy is then derived using dynamic programming. Furthermore, to deal with the large-scale auction volume and campaign budget, we proposed neural network models to fit the differential of the values between two consecutive states. Experimental results on two real-world large-scale datasets and online A/B test demonstrated the superiority of our RLB solutions over several strong baselines and state-of-the-art methods, as well as their high efficiency to handle large-scale data.

For future work, we will investigate model-free approaches such as Q-learning and policy gradient methods to unify utility estimation, bid landscape forecasting and bid optimization into a single optimization framework and handle the highly dynamic environment.
Also, since RLB naturally tackles the problem of budget over- or under-spending across the campaign lifetime, we will compare our RLB solutions with the explicit budget pacing techniques \cite{lee2013real,xu2015smart}.


\section*{ACKNOWLEDGMENTS}
We sincerely thank the engineers from YOYI DSP to provide us the offline experiment dataset and the engineers from Vlion DSP to help us conduct online A/B tests.

{
\bibliographystyle{abbrv}
\bibliography{wsdm306-cai}
}
\end{document}